\documentclass[final]{cvpr}

\usepackage{times}
\usepackage{epsfig}
\usepackage{graphicx}
\usepackage{amsmath}
\usepackage{amssymb}
\usepackage{tabularx}
    \newcolumntype{L}{>{\raggedright\arraybackslash}X}

\usepackage{comment}
\usepackage[dvipsnames]{xcolor}
\newcommand{\smallsec}[1]{\vspace{0.05in} \noindent {\bf #1:}}

\newcommand{\sunnie}[1]{{\color{RedOrange} Sunnie: #1}}
\newcommand{\vivien}[1]{{\color{Lavender} Vivien: #1}}
\usepackage{enumitem}
\usepackage{multirow}
\usepackage{tikzit}
\usepackage[export]{adjustbox}
\usepackage{relsize}
\usepackage{subcaption}

\tikzstyle{circle}=[fill=white, draw=black, shape=circle]

\usepackage[pagebackref=true,breaklinks=true,colorlinks,bookmarks=false]{hyperref}

\def\cvprPaperID{26} 
\def\confYear{CVPR 2021}

\title{Cleaning and Structuring the Label Space of the iMet Collection 2020}

\author{Vivien Nguyen\thanks{Equal contribution. A shorter version of this work was accepted to the CVPR 2021 The Eight Workshop on Fine-Grained Visual Categorization.} \ and Sunnie S. Y. Kim\footnotemark[1] \\
Princeton University\\
{\tt\small \{vivienn, sunniesuhyoung\}@princeton.edu}
}

\begin{document}

\maketitle

\begin{abstract}
The iMet 2020~\cite{zhang2019imet} dataset is a valuable resource in the space of fine-grained art attribution recognition, but we believe it has yet to reach its true potential. We document the unique properties of the dataset and observe that many of the attribute labels are noisy, more than is implied by the dataset description. Oftentimes, there are also semantic relationships between the labels (e.g., identical, mutual exclusion, subsumption, overlap with uncertainty) which we believe are underutilized. We propose an approach to cleaning and structuring the iMet 2020 labels, and discuss the implications and value of doing so. Further, we demonstrate the benefits of our proposed approach through several experiments. Our code and cleaned labels are available at \url{https://github.com/sunniesuhyoung/iMet2020cleaned}.
\end{abstract}


\section{Introduction}

Fine-grained art attribute recognition is a novel area of research within fine-grained visual categorization (FGVC) that concerns images with low inter-class variation and high intra-class variation. It is an important area that can assist museums create and maintain their artwork collections. Since there is a limited number of museum experts, if a visual recognition system can automatically label new images with relevant attributes and optionally have the experts verify the labels, we can likely reduce costs of the cataloging process by a great amount. It can also enable searches for visually related artworks and serve as a useful tool for art history and cultural heritage studies. 

The iMet Collection~\cite{zhang2019imet} is a dataset for fine-grained art attribute recognition introduced in the 6th FGVC Workshop at CVPR 2019. It is the first high-quality artwork image dataset with professional photographs of artworks from The Metropolitan Museum of Art (The Met) and research-grade attribute labels curated or verified by experts. However, while there have been two iMet competitions and another in progress, we were able to find surprisingly little discussion on the dataset, task, or proposed solutions.

In this work, we study the unique challenges of fine-grained art attribute recognition presented by the 2020 version of the iMet Collection dataset (iMet 2020).
We start with identifying the unique properties of the dataset (Section~\ref{sec:properties}). Specifically, we document instances where the annotations are incomplete, inconsistent, or redundant. 
Based on our observations, we argue that we should clean and structure its label space and discuss our motivations (Section~\ref{sec:motivations}). We then propose concrete suggestions to improve the iMet 2020 label space (Section~\ref{sec:approach}).
Finally, we demonstrate the benefits of our proposed changes through several experiments (Sections~\ref{sec:training-experiments} and~\ref{sec:evaluation-experiments}).

\section{Related work}
\label{sec:relatedwork}

\smallsec{Fine-grained art attribute recognition}
Fine-grained art attribute recognition is a novel area of research within FGVC. It is also a relatively new area in artwork recognition where most previous works in the space focus on object retrieval in paintings~\cite{Crowley14a}, 3D-based content retrieval in artworks~\cite{Gorisse_2007_3dretrieval}, or artistic style classification~\cite{pmlr_lecoutre17a,lombardi_style}. 
Recently, many museums have made their collection available online.\footnote{https://www.artnome.com/art-data} Most don't include images or attribute labels, however, so they are not suitable for art attribute recognition.

Two datasets similar to iMet 2020 are the Behance Artistic Media (BAM)~\cite{Wilber_2017_ICCV} dataset and the WikiArt\footnote{https://wikiart.org} online database. BAM consists of 65 million images of contemporary, mostly non-photorealistic artworks from Behance. WikiArt primarily, but not exclusively, contains paintings, and is similar to iMet 2020 in that it centers academic art and contains several attribute labels, such as country, date, style, and medium. The level of specificity and tags available for each individual artwork also vary drastically.
In contrast to both of these, iMet 2020 contains a smaller but more diverse set of works and objects. iMet 2020 is also annotated with a much larger and more diverse set of attributes compared to either BAM or WikiArt.

\smallsec{Hierarchical classification in cultural heritage}
Belhi et al.~\cite{belhi_aiccsa_2018} argue that a hierarchical classification framework is better suited for the diversity of objects and label availability in cultural heritage. They propose a two-stage framework where they first classify each artwork's type, then further classify attributes such as artist, year, genre, style, and medium for paintings. They demonstrate their framework on the WikiArt, Met, and Rijksmuseum datasets. 
Similarly, we argue that hierarchical classification is more suitable for fine-grained art attribute recognition with the iMet 2020 dataset. However, we note that~\cite{belhi_aiccsa_2018} is concerned with a two-level hierarchy of artwork type and attributes and treats all attributes in the same way, whereas we propose structuring and doing hierarchical classification of the attributes.


\smallsec{Learning with structured label spaces}
Deng et al.~\cite{Deng_2014_ECCV} and Ding et al.~\cite{Ding_2015_ICCV} model label relations with Hierarchy and Exclusion graphs and demonstrate that they improve classification performance.
More recent work by Hu et al.~\cite{Hu_2016_CVPR} models label relations with a Structured Inference Network and report further improvements.
Dhall~\cite{dhall2020learning} studies different methods of injecting label-hierarchy information to visual classifiers and proposes jointly embedding labels and images using embedding models.
We expect to gain similar improvements when we leverage the structured label space of iMet 2020 during learning, and believe developing such methods is a promising direction for future work.


\begin{figure}[t!]
    \centering
    \includegraphics[width=\linewidth]{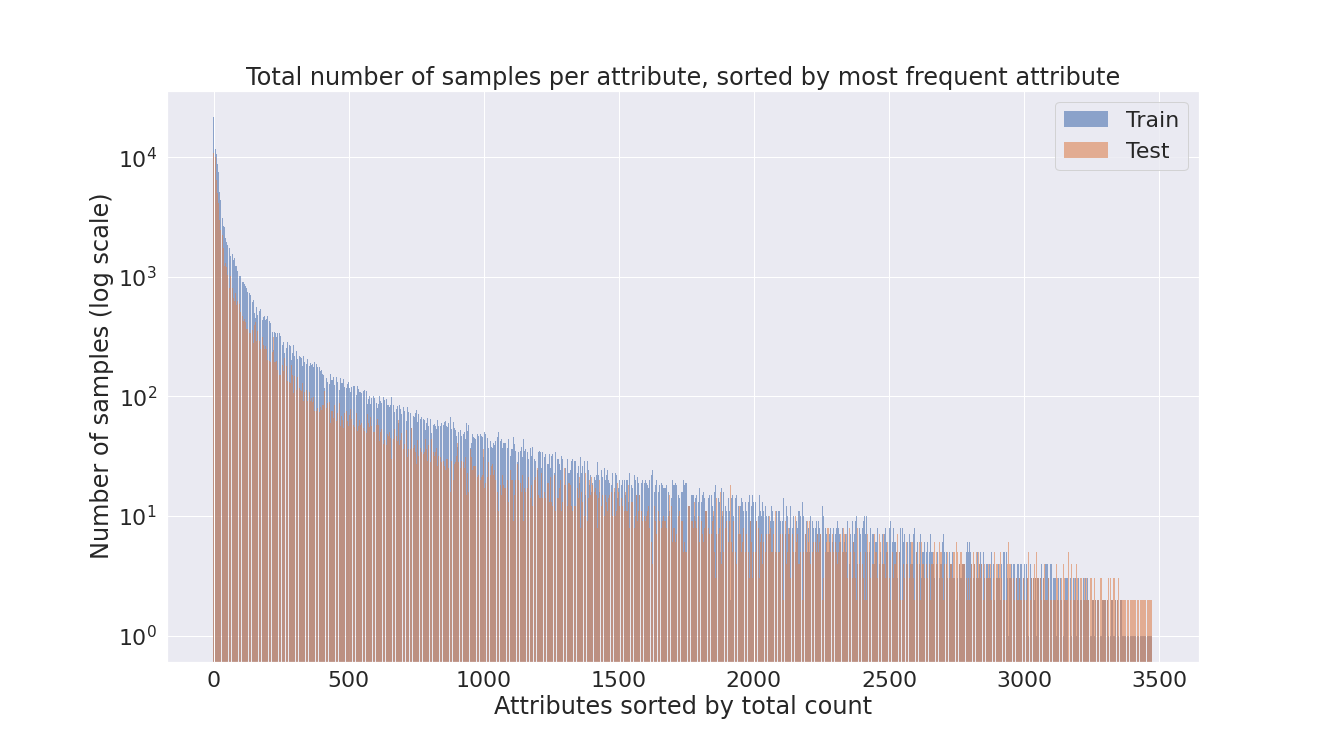}
    \caption{The iMet 2020 dataset has a long-tailed distribution over the 3,474 attributes.}
    \label{fig:attr_counts}
\end{figure}

\begin{figure}[t!]
    \centering
    \includegraphics[width=0.8\linewidth]{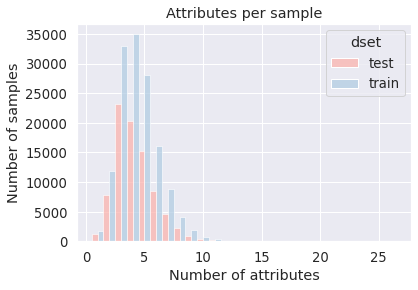}
    \caption{Positive labels are sparse in the iMet 2020 dataset. Although there are 3,474 attributes in the label space, the median number of annotated attributes per sample is 4.}
    \label{fig:attr_len_hist}
\end{figure}

\begin{figure*}[!h]
\caption{Examples of inconsistencies in iMet 2020. While this is a handpicked subset, we find these inconsistencies are quite pervasive throughout the dataset; certainly more than is implied by either the competition or the original dataset paper~\cite{zhang2019imet}.}
\begin{subfigure}[t]{.26\textwidth}
  \centering
     \adjincludegraphics[height=3.9cm, width=.8\linewidth]{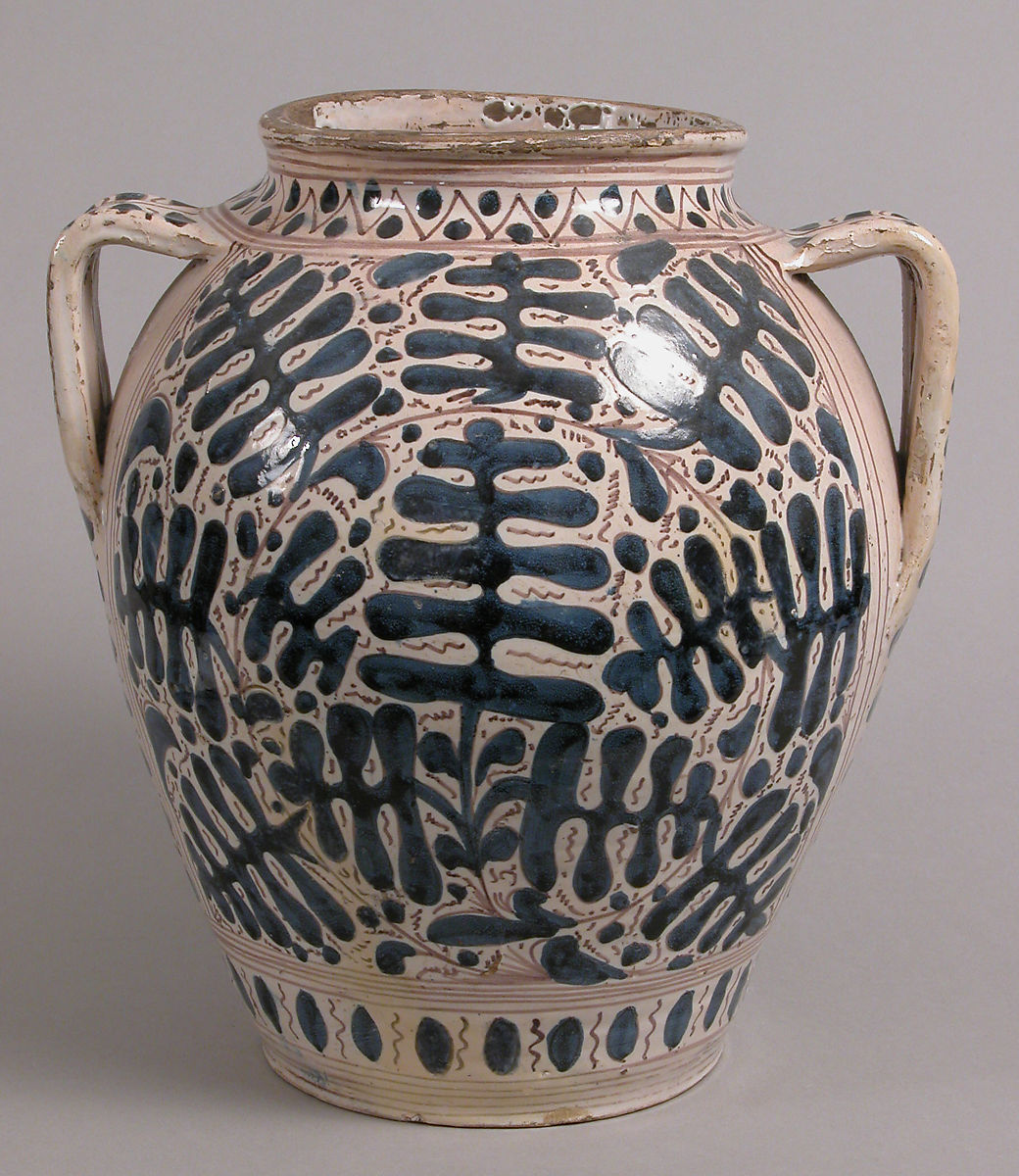} 
  \caption{An early 15th C. Italian jar, provenance listed as Florence, Italy (Central Italy) on The Met's website, but tagged only as {\small\texttt{country:italy}} in iMet 2020.}
  \label{fig:jar}
\end{subfigure}
\hfill
\begin{subfigure}[t]{.26\textwidth}
  \centering
     \adjincludegraphics[height=3.9cm, width=.8\linewidth]{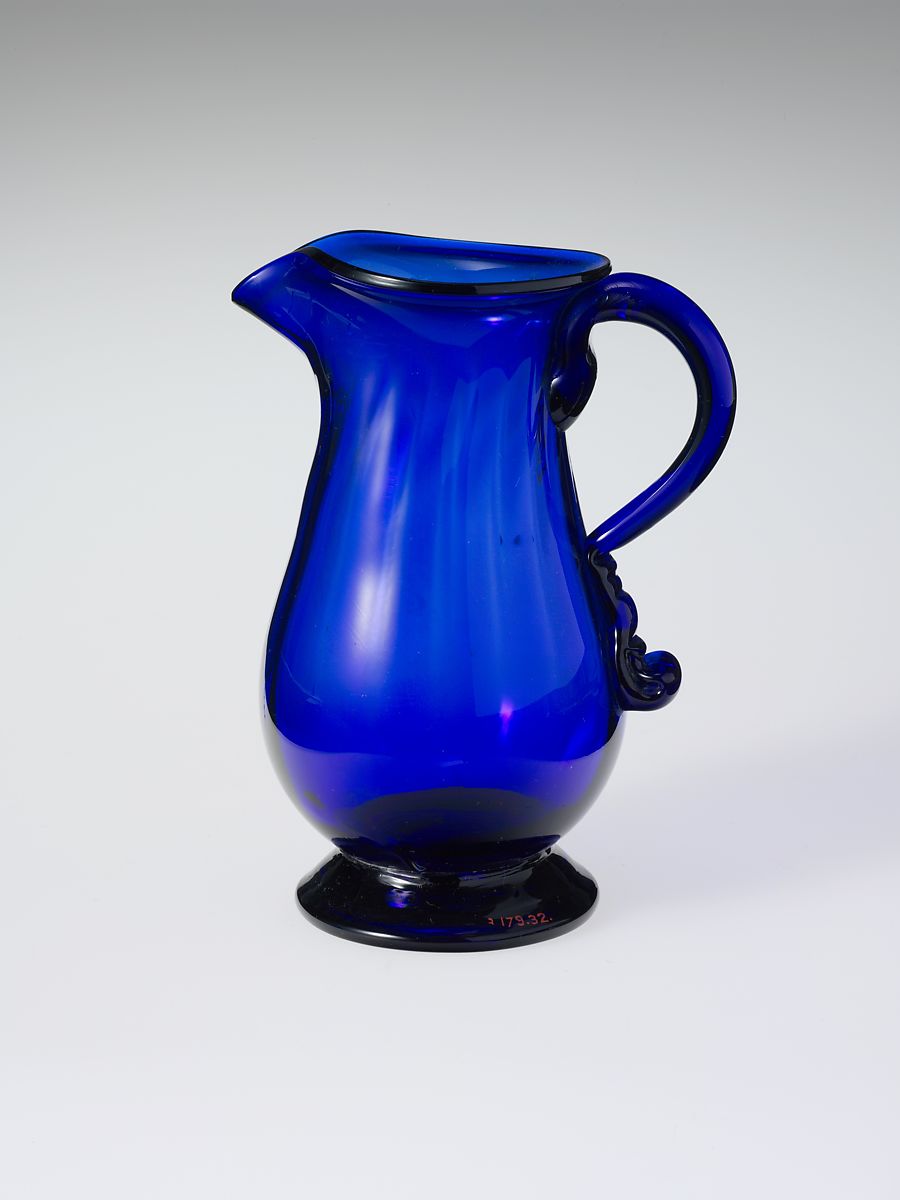}
  \caption{This jug is listed as ``Possibly made in England; Possibly made in United States" on The Met's website but tagged with both countries in iMet 2020.
  }
  \label{fig:jug}
\end{subfigure}
\hfill
\begin{subfigure}[t]{.4\textwidth}
  \centering
\adjincludegraphics[height=3.9cm, width=.8\linewidth]{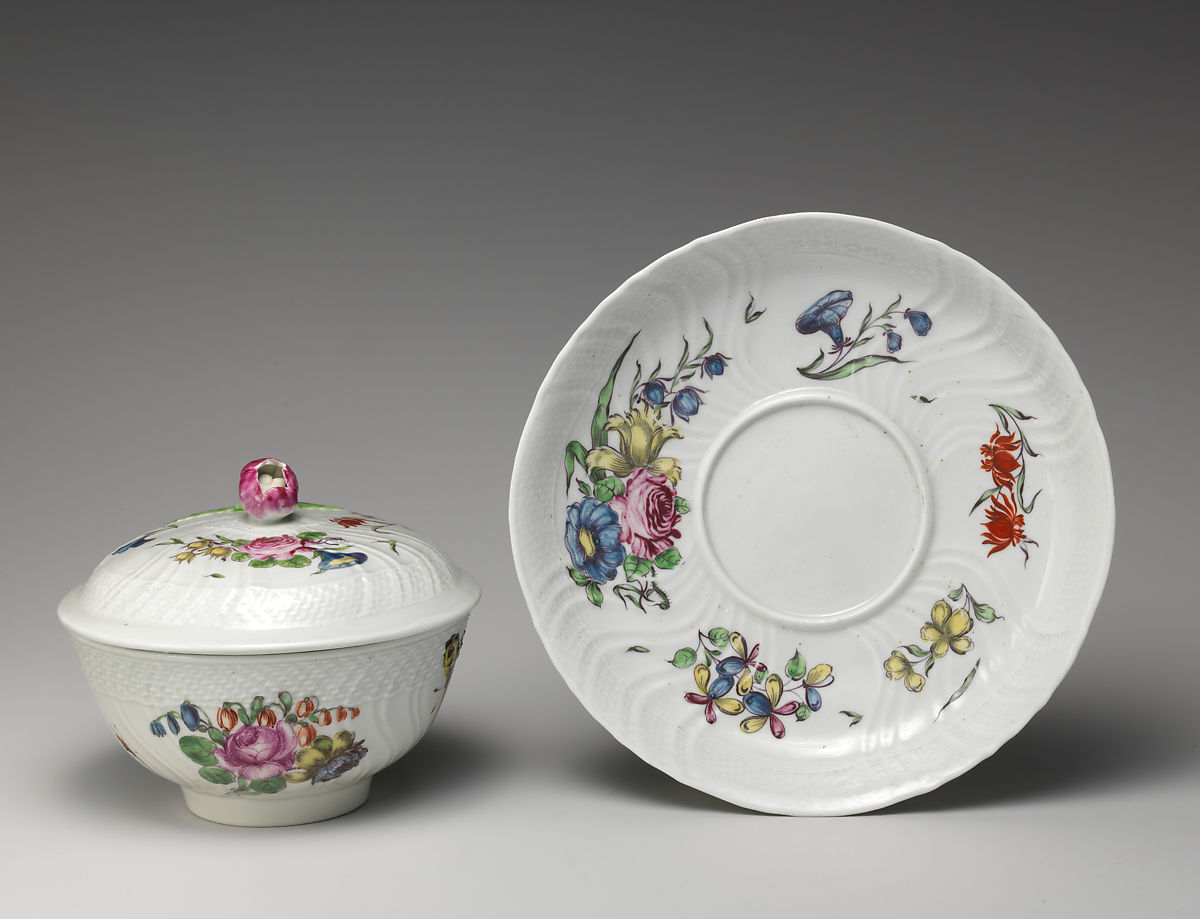}
  \caption{This Worcester porcelain features a variety of flowers, but is only tagged with {\small\texttt{tags:roses}}, while other flower tags exist, as well as a generic catch-all, {\small\texttt{tags:flowers}}.}
  \label{fig:flowers}
\end{subfigure}
\label{fig:examples}
\end{figure*}

\section{Analysis of iMet 2020's label space}
\label{sec:properties}

The iMet 2020 dataset consists of 3,474 binary attribute labels for 226,966 professional photographs of artworks supplied by The Met. The attribute annotations are from subject matter experts or vendor annotators. The annotators were provided with a label taxonomy and access to The Met's online collection, and were instructed to annotate labels which can be visually inferred and to avoid adding labels already present~\cite{zhang2019imet}. 
We highlight that the dataset has a long-tailed distribution over the attributes (Figure~\ref{fig:attr_counts}). Furthermore, positive labels are sparse; the median number of annotated attributes per image is 4 (Figure~\ref{fig:attr_len_hist}).

The 2019 dataset paper \cite{zhang2019imet} describes the iMet Collection as a ``high quality, research grade dataset," while the organizers of the 2020 competition note that participants should consider the annotations noisy.\footnote{https://www.kaggle.com/c/imet-2020-fgvc7/data} We find that both characterizations are true, but we believe the incompleteness and noisiness of the iMet Collection labels inhibit its effectiveness as a dataset for either art historical scholarship or deployable museum systems.

While the attributes that an artwork \textit{is} annotated with are almost always accurate, we have found many instances where the annotations or labels are incomplete, inconsistent, or redundant. 
In this section, we document some concrete examples of these issues for each of the five categories. We also explicitly note other unique or ambiguous aspects of iMet 2020 for future users of the dataset.

\smallsec{country (100 attributes)} First, it is unclear if the attribute label refers to the country where the artwork was (presumably) created, where it was found, or something else. Furthermore, we found that labels vary in specificity (e.g., {\small\texttt{central italy}}, {\small\texttt{present-day greece}}) and some explicitly encode uncertainty (e.g., {\small\texttt{egypt or iraq}}). We discuss inconsistencies surrounding such labels.

For example, there exist labels for both {\small\texttt{italy}} and {\small\texttt{central italy}}. Differentiating the two may be useful: Italy was not a nation-state until 1861, or annotators may want to provide the specific provenance of the work, if known. However, there are inconsistencies between iMet 2020 and The Met's online collection, such as the ``Two-Handled Jar" from Central Italy according to the online collection\footnote{https://www.metmuseum.org/art/collection/search/468166} but only tagged {\small\texttt{italy}} in iMet 2020 (Figure~\ref{fig:jar}).

We interpret that labels such as {\small\texttt{egypt or iraq}} imply the provenance is uncertain, and that experts believe the artwork could be from either country. We note that samples tagged as {\small\texttt{egypt or iraq}} in iMet 2020 are not tagged with the {\small\texttt{egypt}} or {\small\texttt{iraq}}. However, other works with similar uncertainty according to The Met's online collection are tagged with their individual labels to represent multiple possible origins. The ``Cream Jug" described as possibly made in England or the United States in the online collection\footnote{https://www.metmuseum.org/art/collection/search/2601} is tagged both {\small\texttt{england}} and {\small\texttt{united states}} in the dataset (Figure~\ref{fig:jug}).

\smallsec{culture (681 attributes)} Similarly, we found culture attribute labels to have substantial overlap. For example, there are at least 13 attribute labels related to India (e.g., {\small\texttt{india}}, {\small\texttt{india (bengal) or bangladesh}}, {\small\texttt{india (madhya pradesh)}}, {\small\texttt{indian or nepalese}}) with varying levels of specificity and uncertainty. 

As expected, country and culture attribute labels are often correlated. 
This correlation does not imply that the labels are redundant, but for evaluation purposes, it may be useful to represent the relationship between related countries and cultures. We discuss this point further in Section~\ref{sec:motivations}.

\smallsec{dimension (5 attributes)} The five dimension attributes are {\small\texttt{tiny}}, {\small\texttt{small}}, {\small\texttt{medium}}, {\small\texttt{large}}, and {\small\texttt{very large}}. 101,954 of 142,119 images (72\%) are given one of the 5 labels, so they are some of the most common attributes in the dataset. 
The dimension attributes were added in 2020, hence they are not discussed in the 2019 dataset paper~\cite{zhang2019imet}. We were unable to find discussion around the binning of dimension attributes. The Met's website lists exact dimensions of items, while these five attributes are relative labels.

\smallsec{medium (1,920 attributes)} Medium attributes describe the materials utilized in the object's creation and are by far the noisiest. We found more than 100 pairs of attributes that have the same meaning; many are pairs whose spelling differs by one letter such as ({\small\texttt{watercolor}}, {\small\texttt{watercolour}}). 
There are also typos (e.g., ({\small\texttt{commercial lithograph}}, {\small\texttt{commerical lithograph}})) and encoding issues (e.g., ({\small\texttt{copper-gold alloy (shakud\textbackslash x8dō)}}, {\small\texttt{copper-gold alloy (shakudō)}})).

Once again, there are varying levels of specificity and uncertainty among these attributes. For example, there is {\small\texttt{black chalk}}, but also {\small\texttt{black chalk and charcoal}}, {\small\texttt{black chalk on blue paper}}, and {\small\texttt{black chalk or graphite}}. Of the 682 artworks in the training set tagged {\small\texttt{black chalk}}, only one is also tagged {\small\texttt{black}}; of the 6 objects in the training set tagged {\small\texttt{black chalk on blue paper}}, none are tagged with {\small\texttt{black chalk}}.

\smallsec{tags (768 attributes)} Finally, there are tags that cover a wide range of subjects, such as religious icons, historical events, plants and animals. While less overlapping than culture or medium, we still find many hierarchical relationships, such as for named entities ({\small\texttt{goddesses}} as well as {\small\texttt{aphrodite}}), objects ({\small\texttt{flowers}} as well as {\small\texttt{roses}}), and events ({\small\texttt{wars}} as well as  {\small\texttt{Trojan war}}), among others.

While a more general tag such as {\small\texttt{flowers}} could either be used for \textit{all} flowers or \textit{unspecific} flowers, we found that usage is inconsistent (Figure~\ref{fig:flowers}).

\section{Motivations for label cleaning/structuring}
\label{sec:motivations}

Based on our observations in Section~\ref{sec:properties}, we propose two improvements to the labels of iMet 2020. The first is more straightforward: merge identical labels, annotate super/sublabels, ensure that samples are annotated as completely as possible, and develop a consistent encoding of uncertainty. We refer to this as cleaning and completing the iMet 2020 labels. The second is more nuanced: we should organize the label space itself, by providing both hierarchical structure to the labels as well as looser relational annotations. We refer to this as structuring the iMet 2020 labels. In this section, we discuss our motivations for both, then describe our proposed approach in Section~\ref{sec:approach}.

\subsection{Why clean the iMet 2020 labels?} 
Concretely, if we merge identical attributes and annotate supercategories, we can increase the number of samples for many labels. For example, there are 8 training images with the {\small\texttt{bronze gilt}} label and 9 with {\small\texttt{bronze-gilt}}, with no overlap between the images. 
When we merge the two labels, we get more images for {\small\texttt{bronze(-)gilt}}.
Similarly, we can add a supercategory label for the images with its subcategory label (e.g., for images labeled {\small\texttt{black chalk on blue paper}} also label its supercategory, {\small\texttt{black chalk}}). This will increase the number of samples for the supercategory and likely lead to improved recognition performance of it. These simple modifications alleviate some of the challenges associated with label sparsity and long-tailed distributions.

While the existing attribute labels are not incorrect, they are certainly incomplete and noisier than the competition description suggests. Completing and cleaning the iMet labels will reduce inconsistencies, which can provide clearer training signals to models trained on the dataset. Perhaps more importantly, fixing the test set labels will improve iMet's quality as a benchmarking dataset 
~\cite{northcutt2021pervasive,northcutt2021confident}.

Having a complete set of labels could also enable us to apply hierarchy extraction algorithms~\cite{BRUCKER2011724} to automatically determine at least some label relationships.

\subsection{Why structure the iMet 2020 labels?}
\label{sec:why-structure}
Leveraging label relations will lead to better tools for artwork documentation and curation.
After identifying more specific hierarchies beyond the current five categories and structuring the label space, future annotators will be able to annotate a more clean and comprehensive set of attributes for each artwork. For example, an annotator could visually traverse a label hierarchy to easily locate or discover tags.

We believe leveraging semantic relations between labels will lead to more accurate and consistent recognition systems. 
Aside from the benefits from cleaning the iMet labels, modeling label relations can lead to more consistent predictions. For example, we can enforce simultaneous prediction of a supercategory whenever it predicts its subcategory in an image, or enforce mutual exclusion over a set of attributes (such as dimension). Though we usually assume that a recognition model will learn these co-occurrences naturally, previous works~\cite{Deng_2014_ECCV,dhall2020learning,Ding_2015_ICCV,Hu_2016_CVPR} have demonstrated that modeling label relations lead to improvements in classification performance.

Nevertheless, it's possible that a structured label space does not directly yield immediate improvements in the F2 score, which is currently employed as the metric for the iMet competition. In particular, the F2 score prioritizes recall over precision which may incentivize prediction patterns such as predicting multiple dimensions in the ``hopes" of getting one of them right. 

However, using a hierarchical label space itself lends itself not only to different quantitative performance measures, but also more qualitative insights into a model's performance \cite{Kosmopoulos2015}. For example, modeling weak relations between labels (such as the relationship between a geographic country and its corresponding culture) can allow us to select models based on the \textit{kinds} of errors that they make. Intuitively, some errors are more wrong than others, but that goes uncaptured in a flat label space where every label is given equal weight, relative to the others.

We argue that a diverse set of performance measures is necessary to meaningfully evaluate a FGVC system.

\section{Proposed approach}
\label{sec:approach}

We propose the following cleaning and structuring of the iMet 2020 labels. Our goal is not to learn a hierarchy from existing labels (which would only be possible after cleaning) but to correct the inconsistencies and propose a general approach for handling likewise noisy label spaces in FGVC.

\smallsec{Identify and merge identical attributes}
As noted before, we found more than 100 pairs of attributes that have the same meaning in the medium category alone. We found these pairs through fuzzy string matching and manual verification. A relatively naive strategy of calculating the similarity ratio, based on Levenshtein edit distance, of all pairs of attributes worked well.
In Section~\ref{sec:identicalexperiments} we demonstrate performance gains from the merging.

\smallsec{Identify super/subcategories and label supercategories in the presence of subcategories}
We can also use string matching supplemented with manual verification to identify supercategories (e.g., {\small\texttt{black}}) and subcategories (e.g., {\small\texttt{black ink}}, {\small\texttt{black chalk}}). Note that the subcategories may continue branching into more specific attributes, and also that hierarchy need not form a tree. Both {\small\texttt{black}} and {\small\texttt{chalk}} could be considered parents of {\small\texttt{black chalk}}. 
In Section~\ref{sec:supercategoryexperiments} we demonstrate that labeling the supercategory can provide a helpful training signal.

\smallsec{Encode \emph{and} relationships as separate labels}
334 of 3,474 attributes have ``and" in their names. We argue these attributes should be consistently separated into two (or more) labels. That is, {\small\texttt{wool and silk}} should be separated into {\small\texttt{wool}} and {\small\texttt{silk}}. We differentiate this from the above case, where a supercategory subsumes the subcategory.
In Section~\ref{sec:andexperiments} we show that encoding \emph{and} relationships as separate labels yields performance improvements.

\smallsec{Encode \emph{or} relationships}
117 of 3,474 attributes have ``or" in their names. To the best of our knowledge, there is no established way of encoding \emph{or} relationships. One possibility is to label each attribute separately, but encode the \emph{or} relationship in a label relations graph or some other structure so that we can apply a special loss function. For example, for {\small\texttt{french or spanish}}, we can treat a model's prediction correct when it has at least one of {\small\texttt{french}} and {\small\texttt{spanish}}, and incorrect when it has neither.
In Section~\ref{sec:orexperiments} we demonstrate this simple evaluation scheme change.

\smallsec{Encode mutually exclusive relationships}
Some attributes, such as the five dimensions attributes, are mutually exclusive in nature. We believe mutual exclusion is an important semantic relationship that should be encoded, such that we can get valid predictions.
In Section~\ref{sec:dimensionexperiments}, we conduct a brief case study with the dimensions attributes.

\smallsec{Use hierarchical performance measures} 
Desirable properties of performance measures (PMs) for hierarchical classification are outlined in~\cite{10.1007/11766247_34}, and hierarchical PMs are compared to flat performance measures further in~\cite{costa2007review,10.1007/978-3-642-23851-2_59}. PMs based on distance, depth, semantics, or hierarchy are known to provide more discriminative power between types of errors.
Based on these insights, we explore graph-based performance measures in Section~\ref{sec:graphbased-measures}.

\section{Experiments at training time}
\label{sec:training-experiments}

We conduct several proof-of-concept experiments to demonstrate the benefits of our proposed changes. We describe our experimental setup in Section~\ref{sec:experimental-setup}, and the baseline model and its performance in Section~\ref{sec:baseline}. We then discuss experiments that involve altering the dataset and/or the training procedure in Sections~\ref{sec:identicalexperiments}--~\ref{sec:dimensionexperiments}.

\subsection{Experimental setup}
\label{sec:experimental-setup}

\smallsec{Dataset}
We use the iMet 2020 dataset which contains 3,474 attribute labels for 226,966 images (142,119 for training and 84,847 for testing). We do a random 80-20 split of the original training set and use the 80 split as our training set (113,694 images) and the 20 split as our validation set (28,425 images). We only use the test set to report final results. We note that we use a smaller version of the test set (25,958 images) based on the sample\_submission.csv file from the iMet 2020 competition.

\smallsec{Implementation and training details}
For all attribute classifiers, we use ResNet-50~\cite{He2015} pre-trained on ImageNet~\cite{imagenet_cvpr09} classification as the base architecture. We train them using the Adam optimizer, a batch size of 200, and a learning rate of 1e-4 with standard step decay for up to 25 epochs. We use the binary cross entropy (BCE) loss to train all models, unless noted otherwise.\footnote{Our implementation is inspired by Yandex Praktikum's code shared in the competition site: \url{https://www.kaggle.com/alimbekovkz/yandex-praktikum-pytorch-train-baseline-lb-0-699}.}

\smallsec{Evaluation details}
We use 0.1 as the decision threshold for all classes to make the final predictions.
Our main evaluation metric is the micro-averaged F2 score, the official evaluation metric for the iMet competitions. 

\subsection{Baseline model}
\label{sec:baseline}

\begin{table}[t!]
\caption{Analysis of the baseline model. When applicable, we repeat the experiment three times with different random seeds, and report the mean and its standard error.}
\centering
\begin{tabular}{|l|c|} 
\hline
Metric & Value \\
\hline
Micro-averaged F2 score & 64.96 $\pm$ 0.13 \\
Macro-averaged F2 score & 18.65 $\pm$ 0.55 \\
\hline
\# of classes with NaN F2 score & 48 \\
\# of classes with zero F2 score & 1857 \\
\# of classes with positive F2 score & 1569 \\
\hline
Micro-averaged accuracy & 99.85 $\pm$ 0.00 \\
\hline
Overall deviation of F2 score & 28.33 \\
Per-class deviation of F2 score & 4.14 \\
\hline
\end{tabular}
\label{tab:baseline}
\end{table}

Our baseline model is a multi-label attribute classifier trained with the BCE loss on the unmodified training set with 3,474 classes (attributes). We train it three times with different random seeds. Table~\ref{tab:baseline} summarizes the results.

\begin{figure}[t!]
    \caption{Histograms of all (left) and positive-only (right) per-class F2 scores of the baseline model.}
    \centering
    \includegraphics[width=\linewidth]{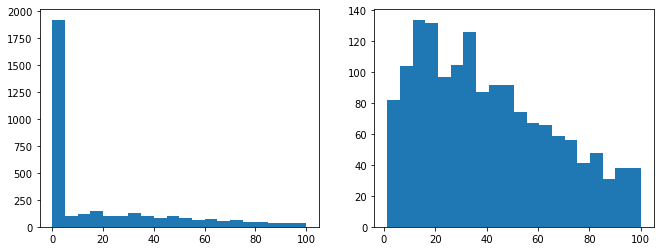}
    \label{fig:hist_perclass_f2}
\end{figure}

Our baseline model achieves a micro-averaged F2 score of 64.96. Note that this score is much higher than the macro-averaged F2 score of 18.65. This difference is explained by the high number of classes with zero or NaN F2 scores and the histograms of per-class F2 scores in Figure~\ref{fig:hist_perclass_f2} which show that the majority of the classes have low F2 scores.
Still, our model is able to achieve a near-perfect accuracy of 99.85 because labels are sparse. Nonetheless, accuracy is neither a robust nor a discriminative metric; an all-negative prediction achieves an even higher accuracy of 99.86. We only report micro-averaged accuracy to highlight the extreme sparsity of positive labels.

Following the suggestion by Gwilliam et al.~\cite{Gwilliam_2021_WACV}, whose work quantifies variance in FGVC results, we also calculate the overall deviation (i.e. standard deviation of the 3$\times$3,474 per-class F2 scores) and the per-class deviation (i.e. standard deviation of per-class F2 scores across 3 runs, averaged over all classes). Consistent with the trend shown in the Figure~\ref{fig:hist_perclass_f2} histograms, the overall deviation is high (28.33). The per-class deviation of 4.14 also suggest that per-class F2 scores fluctuate quite a bit across training runs.

\subsection{Merge of identical attributes}
\label{sec:identicalexperiments}

In this section, we explore the effect of merging identical attributes. In each experiment, we merge {\small\texttt{watercolor}} and {\small\texttt{watercolour}}, {\small\texttt{emerald}} and {\small\texttt{emeralds}}, {\small\texttt{garnet}} and {\small\texttt{garnets}}, and train a new multi-label classifier that has one less attribute than the baseline (i.e. 3473 attributes). In Table~\ref{tab:identical}, we report the number of positive training images, accuracy, and F2 score for each attribute pair before and after the merge. We note that ``before" refers to the baseline model and ``after" refers to the new model, and that all models have comparable overall F2 score of 64.80--66.43.

We see that accuracy is not a discriminative metric for these pairs because the true negatives dominate in all cases; however, the F2 scores show that merging the identical attributes leads to improved performance for all three pairs. Hence, we conclude that, as expected, merging identical attributes is a simple yet effective strategy to gain performance improvements and a cleaner label space.

\begin{table}[t!]
\caption{We report the number of positive training images and F2 score (mean and standard error across three runs) before and after merging identical attributes. Modifications maintain or improve the F2 score for all three attribute pairs.}
\centering
\resizebox{0.99\linewidth}{!}{
\begin{tabular}{|c|c|c|c|c|} 
\hline
\multirow{2}{*}{Attribute} & \multicolumn{2}{|c|}{Num of images} & \multicolumn{2}{|c|}{F2 score} \\ 
\cline{2-5}
 & Before & After & Before & After \\ 
\hline
\texttt{watercolor} & 2094 & \multirow{2}{*}{2102} & \multirow{2}{*}{82.42 $\pm$ 0.82} & \multirow{2}{*}{83.00 $\pm$ 0.60} \\ 
\texttt{watercolour} & 8 & & & \\
\hline
\texttt{emerald} & 21 & \multirow{2}{*}{57} & \multirow{2}{*}{4.07 $\pm$ 5.75} & \multirow{2}{*}{13.38 $\pm$ 10.28} \\ 
\texttt{emeralds} & 36 & & & \\
\hline
\texttt{garnet} & 62 & \multirow{2}{*}{96} & \multirow{2}{*}{16.33 $\pm$ 2.11} & \multirow{2}{*}{21.41 $\pm$ 3.33} \\ 
\texttt{garnets} & 34 & & & \\
\hline
\end{tabular}
}
\label{tab:identical}
\end{table}

\subsection{Labeling of supercategory attributes}
\label{sec:supercategoryexperiments}

Next, we explore the effect of labeling supercategory attributes through one example set of super- and subcategories. In this case study, we treat {\small\texttt{black}} as the supercategory and 83 other attributes with {\small\texttt{black}} in their names (e.g., {\small\texttt{black chalk}}, {\small\texttt{almost black}}) as the subcategories.
Whenever a subcategory is labeled, we also label the supercategory. This modification increases the number of samples with {\small\texttt{black}} in the training set from 169 to 3767.

We train a model on this modified training set, and in Table~\ref{tab:supercategory}, compare it to the baseline on all 3474 attributes, the supercategory {\small\texttt{black}}, and the 83 subcategories. Note that we evaluate the baseline on the unmodified test set and the new model on the modified test set so that the training and test distributions are consistent in each case.
We see that both models perform similarly overall, but the new model performs better on the supercategory and the subcategories. 
The results suggest that the simple modification of labeling the supercategoy provides a more helpful training signal.

\begin{table}[t!]
\caption{We report F2 score (mean and standard error across three runs) for all attributes, the {\small\texttt{black}} supercategory, and 83 subcategories before and after labeling the supercategory attribute. Modification improves the F2 score on all sets.}
\centering
\begin{tabular}{|c|c|c|c|} 
\hline
Classes & Baseline & New model \\ 
\hline
All 3,474 & 64.96 $\pm$ 0.13 &  65.09 $\pm$ 0.23 \\
1 supercategory & 39.69 $\pm$ 5.18 & 77.62 $\pm$ 1.01 \\
83 subcategories & 65.80 $\pm$ 0.92 & 67.10 $\pm$ 0.61 \\
\hline
\end{tabular}
\label{tab:supercategory}
\end{table}

\subsection{Encoding of \emph{and} relationships}
\label{sec:andexperiments}

We also experiment with encoding \emph{and} relationships as separate labels. For the 334 attributes that have \emph{and} in their names, we separate their names into smaller tokens using \emph{and} as the separator. For 170 attributes, all of their smaller tokens are existing attributes (e.g., {\small\texttt{german and italian}} is split into {\small\texttt{german}} and {\small\texttt{italian}} both of which are attributes in iMet 2020). For 30 attributes, none of their smaller tokens are existing attributes (e.g., {\small\texttt{weights and measures}}). For the remaining 134 attributes, only a part of their smaller tokens are existing attributes.

Our modification is simple. Whenever a smaller token is an existing attribute, we label that attribute. To reduce redundancy, we remove the 170 attributes whose smaller tokens are all existing attributes. We then train a model with $3,474-170=3,304$ attributes. This model achieves a micro-averaged F2 score of 66.03 $\pm$ 0.50, higher than the baseline's 64.96 $\pm$ 0.13, suggesting that encoding \emph{and} relationships as separate labels is useful in practice.


\subsection{Mutual exclusion of dimension attributes}
\label{sec:dimensionexperiments}

Next, we explore the effects of utilizing the knowledge that dimensions are mutually exclusive. We don't modify any of the labels, since the dimension attributes are already exclusive (or unlabeled).
We do make a change to how the attribute classifier is trained. Specifically, we enforce mutual exclusion by applying the softmax operator across the five dimensions attributes before computing the BCE loss. We call this new classifier the ``exclusive dim model."

For clarity, we restrict our analysis to (1) only the dimension attributes, and (2) only the test samples for which a ground truth dimension label exists. We analyze three sets of predictions: (1) predictions from the baseline model where scores are thresholded at 0.1 (such that multiple or no predictions are also possible), (2) predictions from the baseline model where the top score is picked (such that the model will always predict exactly one dimension), and (3) predictions from the exclusive dimensions model where the top  score is picked (again, such that the model will always predict exactly one dimension). Accuracy and F2 scores for these predictions are shown in Table~\ref{tab:excl_dim}.

The difference between accuracy and F2 score for the baseline model suggests that \emph{not} restricting the model does give some benefit. The baseline model seems to be able to pick up some extra correct predictions if not restricted to only one. However, all three models perform about the same when considering accuracy.
This insight can help researchers tune models along a particular attribute vertical.

\begin{table}[t!]
\caption{Results on enforcing mutual exclusion among dimensions attributes. We report accuracy and F2 score. Baseline model is the best performing according to the F2 score, but enforcing a single prediction (from the baseline model) yields the best accuracy.}
\centering
\resizebox{\linewidth}{!}{
\begin{tabular}{|c|c|c|} 
\hline
Prediction type & Accuracy & F2 score \\ \hline
Baseline model & 43.16 $\pm$ 0.37 &  72.23 $\pm$ 0.06 \\ \hline
Baseline, top dim only & 42.58 $\pm$ 0.26 & 59.27 $\pm$ 0.36 \\ \hline
Exclusive dim model & 42.49 $\pm$ 0.17 & 59.18 $\pm$ 0.22 \\ \hline
\end{tabular}
}
\label{tab:excl_dim}
\end{table}

\section{Experiments at evaluation time}
\label{sec:evaluation-experiments}

While we hope that incorporating hierarchical knowledge/relationships can improve any model, realistically, many models \emph{suffer} in performance after incorporating knowledge graphs due to the ``preferences" of flat performance measures. Thus, it's important to also consider alternate evaluation metrics. 

Furthermore, it's possible that structuring the label space only at evaluation time, without changing the training of the model, can give us new measures to gain new insights into model performance. In this section, we explore such experiments and interpret how comparing these measures with the flat F2-score can give further intuition into where the model is making errors.

\subsection{Encoding of \emph{or} relationships}
\label{sec:orexperiments}

Similar to how we analyzed \emph{and} relationships, for the 117 attributes that have \emph{or} in their names, we separate their names into smaller tokens using \emph{or} as the separator.
For 77 attributes, all of their smaller tokens are existing attribues (e.g., {\small\texttt{british or irish}}).
For 11 attributes, none of their smaller tokens are existing attributes (e.g., {\small\texttt{spanish or mexican}}).
For the remaining 29 attributes, only a part of their smaller tokens are existing attributes.
We note that most \emph{or} attributes are country or culture attributes.

We explore changing the evaluation scheme for \emph{or} relationships. That is, for the attribute {\small\texttt{british or irish}}, we consider a prediction is correct if it includes one of {\small\texttt{british or irish}}, {\small\texttt{british}}, or {\small\texttt{irish}}. For the baseline model, this evaluation scheme changes the F2 score of the 117 \emph{or} attributes from 43.35 $\pm$ 1.41 to 73.45 $\pm$ 0.28. 

Naturally, these scores are not directly comparable; for one, we only alter the way true positives are counted, and not false positives or false negatives. However, the increase in score itself provides valuable insight into the model's performance. In particular, this indicates that the model is predicting many related attributes, even if not correctly predicting the \emph{or} attribute. Depending on the desired results, researchers can re-label samples, re-encode uncertainty, or focus on teaching the model to learn to discriminate between certain/uncertain samples.

\subsection{Using graph-based performance measures}
\label{sec:graphbased-measures}

As described in Section~\ref{sec:why-structure}, intuitively, some errors are better (or worse) than others. In this section, we describe a metric that gives ``partial credit" to models for predicting ``reasonable" (but wrong) labels.

\begin{figure}
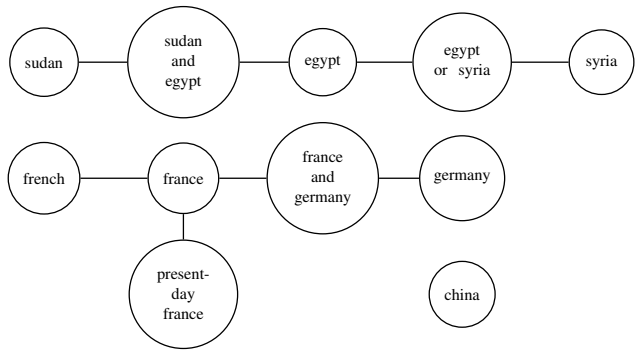

\caption{Subset of our country label relations graph. In this subgraph, {\small\texttt{china}} has no graph neighbors, so no partial credit can be given for any answer except the exact true label. On the other hand, a path exists between {\small\texttt{french}} and {\small\texttt{present-day france}} via {\small\texttt{france}}, and so have some notion of ``similarity".}
\ctikzfig{countries-graph}
\label{fig:graph}
\end{figure}

We focus on country attributes and construct a graph to model connections between existing attributes. To make the process as straightforward as possible, we focus on attributes involving \emph{and} or \emph{or} connections, and attributes with ``obvious" relationships. For example, consider the label {\small\texttt{sudan and egypt}}: since {\small\texttt{sudan}} and {\small\texttt{egypt}} already exist in the label set, we can draw an undirected edge between each of those labels and {\small\texttt{sudan and egypt}}. An example of an `obvious" relationship between labels {\small\texttt{french}} and {\small\texttt{france}}. We also draw edges between ({\small\texttt{united kingdom}}, {\small\texttt{england}}) and ({\small\texttt{united kingdom}}, {\small\texttt{scotland}}). A small subset of the resulting graph can be seen in Figure~\ref{fig:graph}.

Kosmopoulos et al. \cite{Kosmopoulos2015} review and extend several metrics for evaluating hierarchical classification systems. Their work includes multi-label classification problems, but focuses on non-cyclical and explicit hierarchies, i.e. a directed graph where descendants and ancestors can be established for a particular node.

The graph we create is undirected; thus we can't use metrics that evaluate the performance of the model based on the level of specificity the model is able to predict. However, we can simply use the distance in the graph between the true and predicted label as a proxy for similarity. Likewise, we can punish the model less harshly for predicting incorrect but close labels. 

To be precise, we redefine the calculation of true positives (TP), false positives (FP), and false negatives (FN) which are then used to calculate the F2 score.
Let $d(T, P)$ be the distance in the graph between true label $T$ and predicted label $P$. $d(T,P) = 0$ implies that $T == P$, and $d(T,P) = \infty$ implies no path exists between $T$ and $P$.
We compute for each sample:
\begin{align}
TP = \sum_{T}\max_{P}\frac{1}{d(T,P) + 1}\\
FN = \sum_{T}\max_{P}(1 - \frac{1}{d(T,P) + 1})\\
FP = \sum_{P}\frac{1}{d(T,P) + 1}
\end{align}
and calculate the F2 score with these counts. We refer to this metric as the ``graph" F2 score, and compare it to the ``normal" F2 score.

\subsubsection{Comparing metrics}

To compare metrics, we use two notions suggested by Huang and Ling \cite{compare-metrics}, \textbf{consistency} and \textbf{discriminancy}, defined below.
Suppose we want to compare two different performance measures, $f$ and $g$, on their evaluation of two different models, $A$ and $B$.

\smallsec{Definition 1: Consistency} Two measures $f$ and $g$ are \textit{strictly} consistent with one another if there exist no models $A$ and $B$ where $f(A) > f(B)$ but $g(A) < g(B)$. We can also loosen this definition to establish a \textit{statistical degree} of consistency:
\begin{align}
R & = \{(A, B) | f(A) > f(B), g(A) > g(B)\} \\
S & = \{(A, B) | f(A) > f(B), g(A) <= g(B)\}
\end{align}
\begin{align}
DoC(f, g) & = \frac{|R|}{|R| + |S|} \\
DoC(f,g) & = DoC(g,f)
\end{align}

\smallsec{Definition 2: Discriminancy} A performance measure $f$ is \textit{strictly} more discriminative than a measure $g$ if there exists models $A$ and $B$ for which $f(A) > f(B)$ but $g(A) = g(B)$, and no models $A$ and $B$ for which $g(A) > g(B)$ but $f(A) = f(B)$. Again, we can establish a \textit{statistical degree}\footnote{Huang and Ling \cite{compare-metrics} use a strict equality in these equations.} of discriminancy:
\begin{align}
P & = \{(A, B) | f(A) > f(B), g(A) \approx g(B)\} \\
Q & = \{(A, B) | g(A) > g(B), f(A) \approx f(B)\}
\end{align}
\begin{align}
DoD(f, g) & = \frac{|P|}{|Q|} \\
DoD(f,g) & = \frac{1}{DoD(g,f)}
\end{align}
According to \cite{compare-metrics}, if $DoC(f,g) > .5$ and $DoD(f,g) > 1$, then intuitively, metric $f$ is better than metric $g$.

\subsubsection{Graph-based metrics are experimentally consistent and more discriminative}

To quickly obtain a large number of models to compare the graph F2 score and normal F2 core, we use a single baseline ResNet-50 model (as described in Section~\ref{sec:experimental-setup}) but vary the prediction threshold between [0.0025, 0.5]. 
We then compute that $DoD(graph, normal) = 1.77$ and $DoC(graph, normal) = 0.92$. Thus, both metrics are consistent with one another, and the graph F2 score is slightly more discriminative than the normal F2 score in this setting.

\begin{figure}
\centering
\caption{Parts of the graph that are flat are intervals where that metric has low discriminancy. Particularly around threshold 0.1, we see that the graph metric has improved discriminancy (i.e. a slightly steeper slope).}
\includegraphics[width=\linewidth]{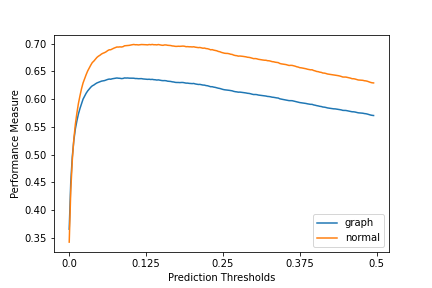}
\label{fig:perf-measure-comp}
\end{figure}

\section{Conclusion}
\label{sec:conclusion}

In this work, we studied the unique challenges of fine-grained art attribute recognition presented by the iMet Collection 2020~\cite{zhang2019imet} dataset. We first documented the unique properties of the dataset for each of the five attribute categories. We then discussed the motivations for cleaning and structuring the label space, and proposed an approach for them. Through several experiments, we also demonstrated the benefits of our proposed changes. We hope our work helps future users of the dataset and practitioners in fine-grained art attribute recognition, and serves as a useful resource in handling other noisy label spaces in FGVC. 

\smallsec{Acknowledgments}
This work was done as a final project for the COS 529 Advanced Computer Vision course at Princeton University. We thank our advisors Szymon Rusinkiewicz and Olga Russakovsky for their support. We also thank members of the UC Berkeley Ng Lab and the Princeton Visual AI Lab for their helpful comments. Finally, we thank the organizers of the iMet Competition, particularly Chenyang Zhang, and the FGVC8 workshop reviewers for suggestions and support.

{\small
\bibliographystyle{ieee_fullname}
\bibliography{egbib}
}

\end{document}